\documentclass{article}

\usepackage{arxiv}
\usepackage[utf8]{inputenc} 
\usepackage[T1]{fontenc}    
\usepackage{hyperref}       
\usepackage{url}            
\usepackage{booktabs}       
\usepackage{amsfonts}       
\usepackage{nicefrac}       
\usepackage{microtype}      
\usepackage{lipsum}         
\usepackage{graphicx}
\usepackage{subcaption}
\usepackage{amsmath}
\usepackage{cleveref}       
\usepackage{pifont}
\usepackage[numbers,sort&compress]{natbib} 
\usepackage{doi}
\usepackage{textcomp}%
\usepackage{xcolor}%

\title{Brain vascular age prediction using cerebral blood flow velocity and machine learning algorithms}
\newif\ifuniqueAffiliation
\uniqueAffiliationtrue

\author{\hspace{1mm}Anni Zhao\\Center for Data Science\\
    Nell Hodgson Woodruff School of Nursing\\
    Emory University\\
    Atlanta, GA 30322, USA\\
    \texttt{anni.zhao@emory.edu}
    \And
    \hspace{1mm}Alex Bateh\\
    Division of Nephrology\\
    Department of Medicine\\
    University of Alabama at Birmingham\\
    Birmingham, AL 35294, USA\\
    \texttt{abateh@uab.edu}
    \And
    Tyler Baldridge\\
    Department of Neurology\\
    School of Medicine\\
    University of Kansas Medical Center\\
    Kansas City, KS 66103, USA\\
    \texttt{tbaldridge@kumc.edu}
    \And
    Sandra Billinger\\
    Department of Neurology\\
    School of Medicine\\
    University of Kansas Medical Center\\
    Kansas City, KS 66103, USA\\
    \texttt{sbillinger@kumc.edu}
    \And
    Xiao Hu\thanks{Corresponding author: Xiao Hu.} \\
    Center for Data Science\\
    Nell Hodgson Woodruff School of Nursing\\
    Emory University\\
    Atlanta, GA 30322, USA\\
    \texttt{xiao.hu@emory.edu}
}

\renewcommand{\shorttitle}{\textit{arXiv} Template}

\begin{document}
\maketitle

\begin{abstract}
Defining vascular age in terms of physiological function has become one focal point of the extensive studies to categorize and track chronological age. Transcranial Doppler (TCD) is a method by which cerebral blood flow velocity is measured along the major arteries feeding the human brain. This study aims to use features extracted from TCD to estimate chronological age and assess accelerated aging in subjects with various brain diseases. We predict subjects with various brain diseases to present with accelerated cerebrovascular aging when tested on various regression models trained by healthy subjects. 168 healthy subjects and 277 diseased subjects with bilateral TCD recordings of the middle cerebral artery were analyzed using the Morphological Analysis and Clustering of Intracranial Pressure (MOCAIP) algorithm. MOCAIP-generated features and heart rate variability features were used as input features for regression models to predict the brain vascular age. 66 subjects with acute stroke, 27 subjects with post stroke, 26 subjects with Alzheimer's disease, 23 subjects with mild cognitive impairment, and 135 established subjects were tested against the machine learning model to assess for accelerated cerebrovascular age. The trained model, on average, predicted healthy subjects' cerebrovascular age to be 3.69 years above their chronological age. Subjects with different disease conditions exhibited varying levels of age acceleration. The differences in healthy and diseased subjects' performances suggest that features generated using TCD may be relevant when evaluating accelerated cerebrovascular aging. Moreover, imbalanced datasets have been observed to affect the performance of machine-learning-based brain age prediction models.
\end{abstract}

\keywords{Brain vascular age prediction \and Cerebral blood flow velocity \and Morphological analysis and clustering of intracranial pressure \and Cerebrovascular aging}

\section{Introduction}\label{sec1}
Brain vascular aging reflects the progressive structural and functional changes in the cerebral vasculature over time, and it is closely associated with cognitive decline, stroke risk, and other neurological disorders. Accurate prediction of brain vascular age can therefore provide a meaningful biomarker for assessing cerebrovascular health beyond chronological age. Furthermore, a reliable machine learning model plays an important role in the assessment of therapeutic outcomes. In recent years, machine learning algorithms have emerged as a powerful tool for brain vascular age prediction because they can capture complex, nonlinear relationships from physiological \cite{LibisellerPhelan2022, ShinNoh2022}, imaging \cite{HanGe2023}, and multimodal datasets \cite{ChattopadhyaySenthilkumar2026}. By learning patterns associated with vascular aging, these methods offer the potential to identify individuals with accelerated vascular aging at an early stage, support personalized risk assessment, and improve preventive interventions. As a result, brain vascular age prediction using machine learning has become a promising direction in precision medicine and computational neuroscience. Various algorithms and morphologies have been adopted for brain age prediction, including convolutional neural network \cite{PoloniFerrari2022}, cortical structure \cite{MadanKensinger2018}, and Hidden Markov Model \cite{WangPham2011}. Furthermore, there are some algorithm specialities adopted for brain age prediction, such as the decentralized algorithm \cite{BasodiRaja2021} and extreme learning machine framework \cite{KassaniGossmann2019}. However, there is a large amount of heterogeneity in model performance reported between studies using machine learning models. Mainly, the machine learning model will be trained first on the labeled data from healthy subjects, and implemented for brain age prediction for patients with brain diseases. A comprehensive review of the brain vascular age prediction using machine learning models from 2013 to 2024 can be found in \cite{KumariSundarrajan2024}. Another review focusing on the UK Biobank for brain age prediction can be found in \cite{LiGao2026}. 

Most of the existing research focused on using MRI-based features for brain vascular age prediction. MRI-based features have shown a significant relationship with the brain vascular age. A well-developed brain vascular age prediction model, DeepBrainNet, has been trained using a large set of MRI scans \cite{BashyamErus2020}. Features extracted from various physiological signals have also been adopted for brain vascular age prediction using machine learning algorithms. Recent studies suggest that age-related vascular and neurophysiological changes can be quantified from noninvasive physiological signals, including transcranial Doppler cerebral blood flow velocity \cite{MillerGhisletta2021}, near-infrared spectroscopy–derived cerebral pulsation, photoplethysmography \cite{NieZhao2025}, and EEG \cite{EngemannMellot2022, HuXiang2025}. These findings support the feasibility of developing machine-learning models to estimate brain vascular age from cerebral hemodynamic signal features. Photoplethysmogram (PPG) was adopted for brain vascular age prediction using deep convolutional neural networks in \cite{ShinNoh2022}. It has been studied that the artificial intelligence-derived photoplethysmography (AI-PPG) age can serve as a biomarker for cardiovascular health \cite{NieZhao2025}. A distribution-aware loos function was first designed to reduce bias from the imbalanced age distribution. However, limited attention has been given to the imbalanced distribution in brain vascular age prediction. In this paper, the cerebral blood flow velocity measured from TCD and the extracted features are adopted for brain vascular age prediction. It has been investigated that cerebral blood flow velocity is an important aging-related physiological signal. The morphological features extracted from the cerebral blood flow velocity are closely related to age and neuropsychological performance \cite{PaseGrima2014, MillerGhisletta2021}. Furthermore, the influence of dataset imbalance has been investigated to highlight the importance of carefully selecting training and testing datasets for brain vascular age prediction.

The paper is organized as follows. Section \ref{sec_datasets} explains the datasets that were adopted for the training of the machine learning model. Section \ref{sec_algorithm} shows the feature extraction and machine learning algorithms. Section \ref{sec_results} shows the results of the brain vascular age prediction using various machine learning algorithms. Section \ref{sec_conclusion} concludes the paper. 

\section{Datasets}
\label{sec_datasets}
\begin{figure*}[ht]
	\centering
	\includegraphics[width=1.0\textwidth]{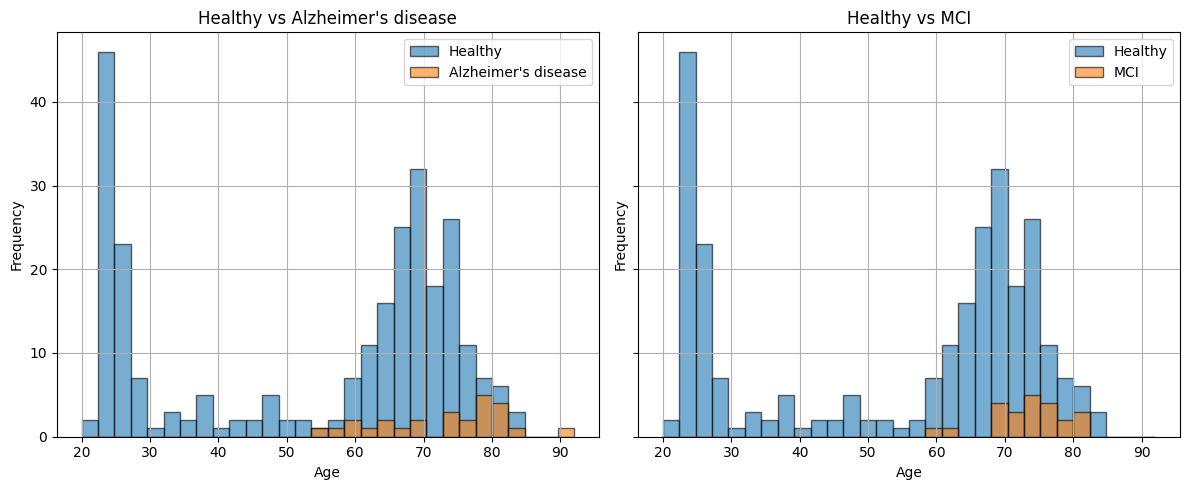} 
	\caption{Data distribution of healthy subjects and diseased subjects with Alzheimer's and MCI diseases.}
	\label{FigZhao1}
\end{figure*}

\begin{figure*}[ht]
	\centering
	\includegraphics[width=1.0\textwidth]{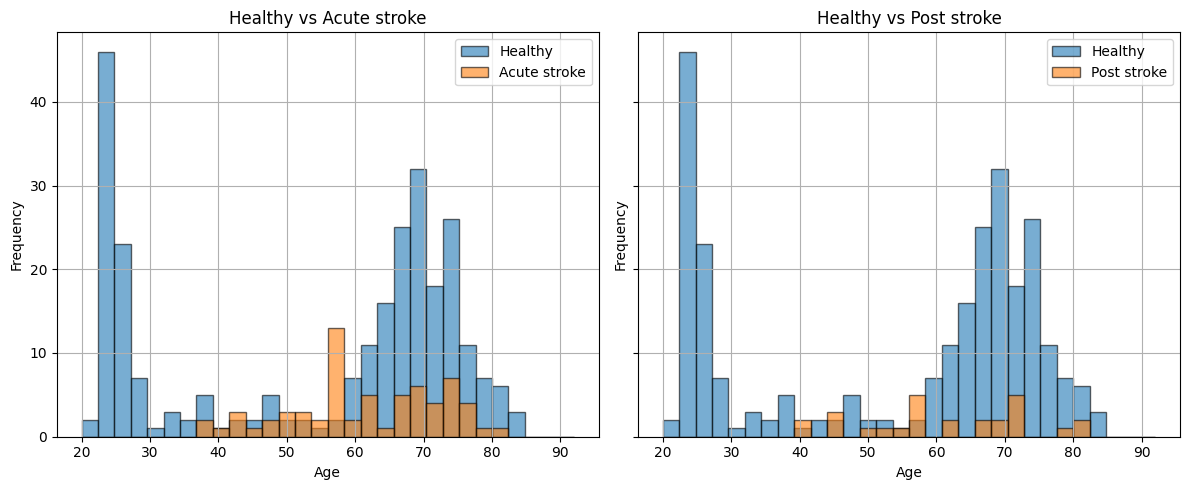} 
	\caption{Data distribution of healthy subjects and diseased subjects with acute stroke and post stroke diseases.}
	\label{FigZhao2}
\end{figure*}

\begin{figure*}[ht]
	\centering
	\includegraphics[width=0.75\textwidth]{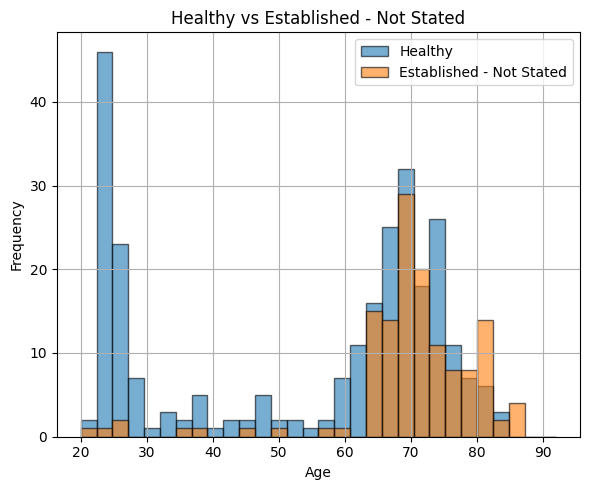} 
	\caption{Data distribution of healthy subjects and established subjects.}
	\label{FigZhao3}
\end{figure*}

Currently, we are mainly using the datasets from the University of Kansas, including 168 healthy subjects and 277 subjects with various diseases. There are 26 patients with Alzheimer's disease (AD), 23 patients with mild cognitive impairment (MCI), 66 patients with acute stroke, 27 patients with post stroke, and 135 patients are established subjects. Established subjects refer to individuals with a family history of brain disease who may have a higher risk of developing brain disease in the future. Currently, the established subjects do not show any apparent pattern associated with brain diseases. The distribution of the healthy subjects and disease subjects are shown from Fig. \ref{FigZhao1} to \ref{FigZhao3}.

\begin{figure*}[ht]
	\centering
	\includegraphics[width=0.85\textwidth]{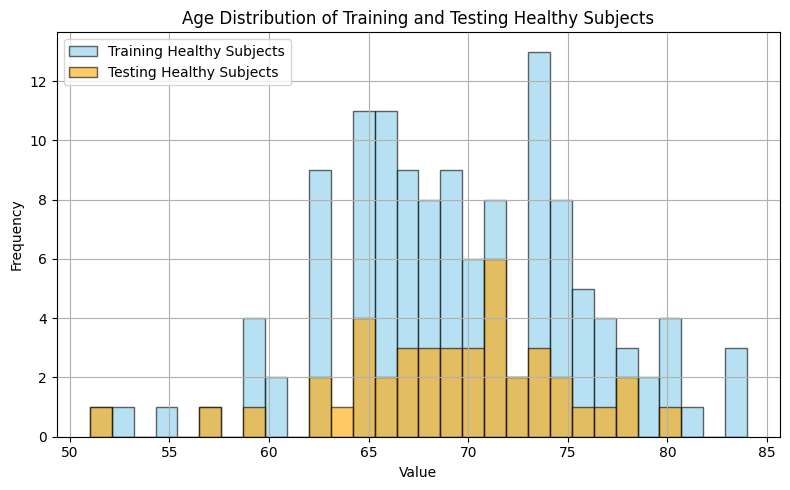} 
	\caption{Data distribution of the training and testing healthy subjects for machine learning algorithms.}
	\label{FigZhao4}
\end{figure*}

As observed from Fig.1, we can see that for healthy subjects, most of the datasets are distributed in the age range [20, 30] and [60, 80], and for various diseased subjects, most of the datasets are distributed above age 50. It has been observed that the age distribution for healthy and diseased subjects is highly imbalanced. In this case, the property of imbalancing may influence the performance of the machine learning model. To further improve model performance and ensure age-distribution consistency, healthy subjects over 50 years old were selected as the training dataset because most diseased subjects were also older than 50. The distribution of the split training and testing healthy subjects is shown in Fig. \ref{FigZhao4}. 75\% of the healthy subjects are adopted as the training datasets, 25\% of the healthy subjects are adopted as the testing datasets.

\section{Algorithm}
\label{sec_algorithm}
In this section, the feature extraction algorithm and machine learning algorithms adopted in this paper are illustrated in detail. The Morphological Clustering and Analysis of the Continuous Intracranial Pressure (MOCAIP) algorithm is adopted to extract the morphological features from the cerebral blood velocity \cite{HuXu2008}. After feature extraction, the extracted features are served as inputs for the machine learning model, and the age is adopted as the output for the model. 

\subsection{Feature extraction algorithm}
\subsubsection{Pre-analysis}
\begin{figure*}[htbp]
    \centering

    \begin{subfigure}{0.48\textwidth}
    	\centering
    	\includegraphics[width=\linewidth]{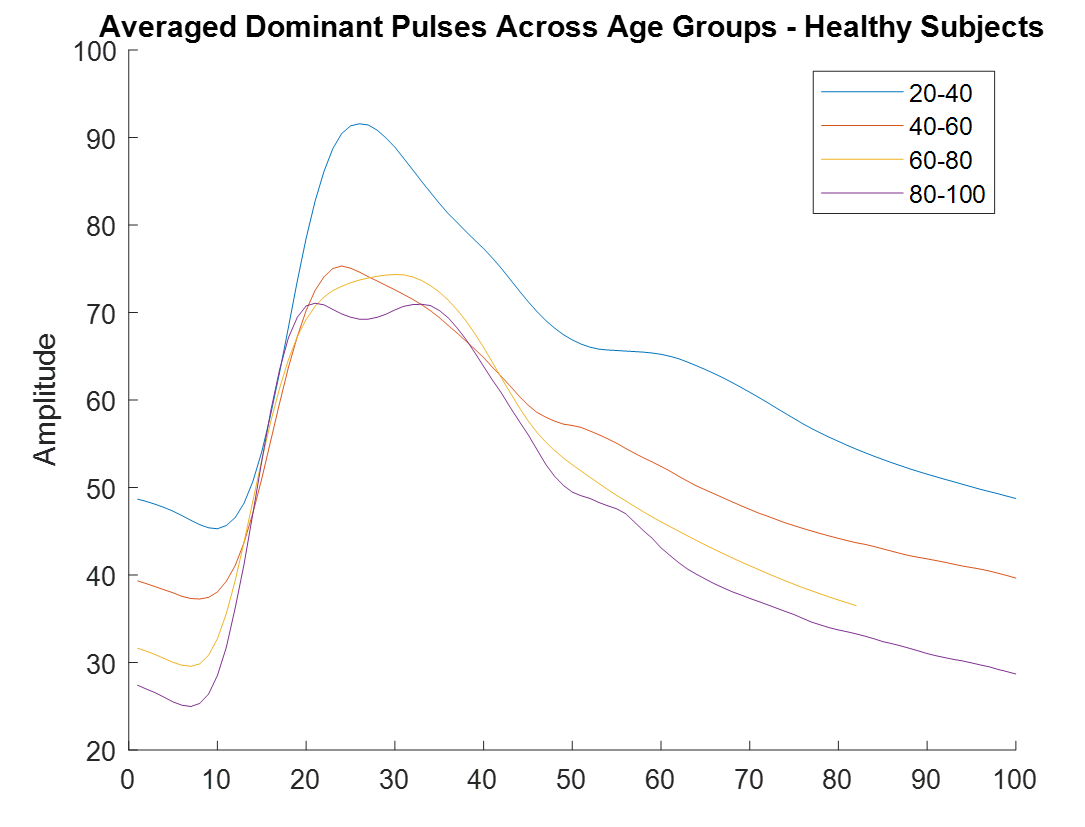}
    	\caption{Averaged dominant pulse comparisons across various age groups for healthy subjects.}
    	\label{FigZhao5}
    \end{subfigure}
    \hfill
    \begin{subfigure}{0.48\textwidth}
        \centering
        \includegraphics[width=\linewidth]{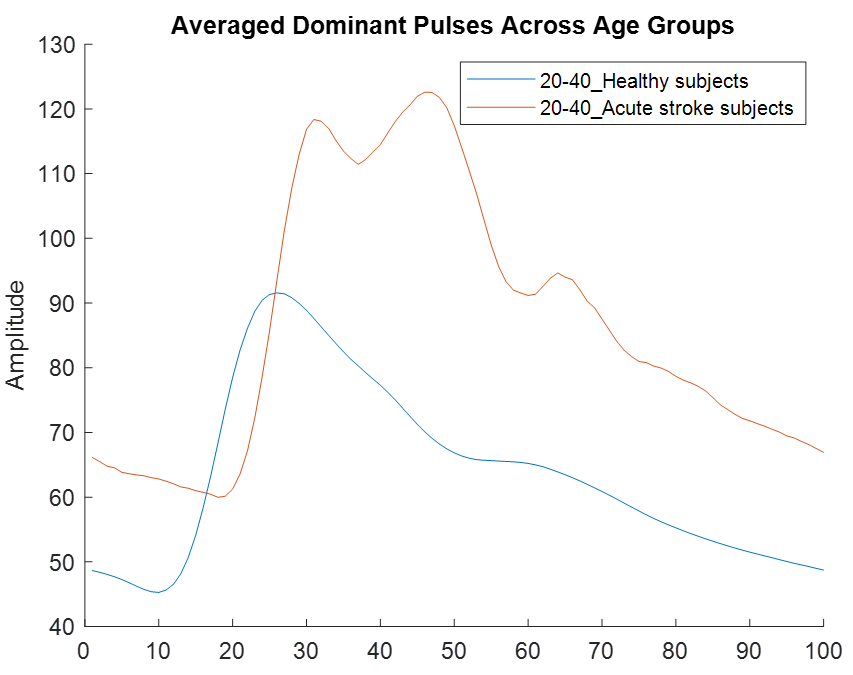}
        \caption{Averaged dominant pulse comparisons in age group [20,40] for healthy and acute stroke subjects.}
        \label{FigZhao6}
    \end{subfigure}

    \vspace{0.3cm}

    \begin{subfigure}{0.48\textwidth}
        \centering
        \includegraphics[width=\linewidth]{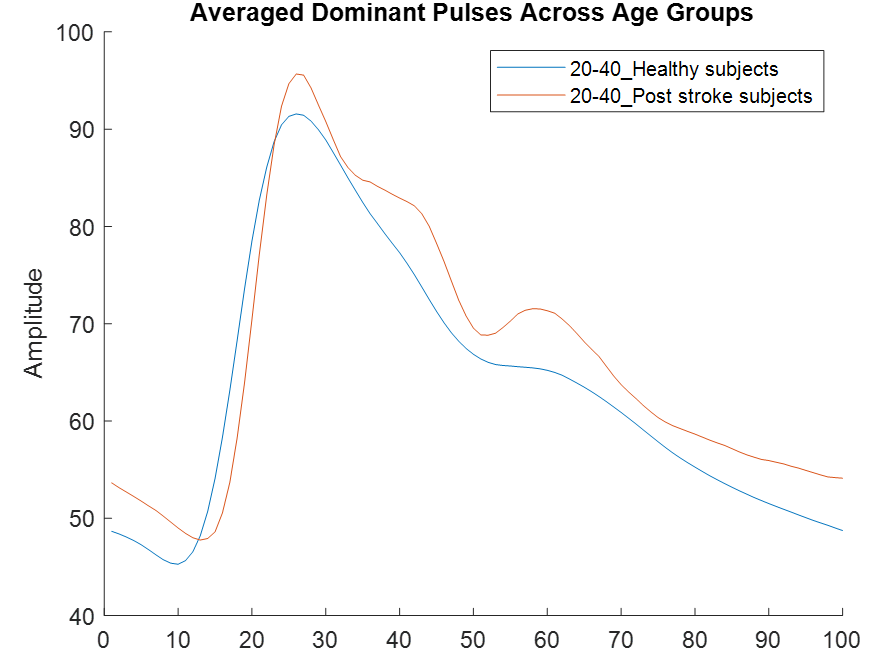}
        \caption{Averaged dominant pulse comparisons in age group [20,40] for healthy and post stroke subjects.}
        \label{FigZhao7}
    \end{subfigure}
    \hfill
    \begin{subfigure}{0.48\textwidth}
        \centering
        \includegraphics[width=\linewidth]{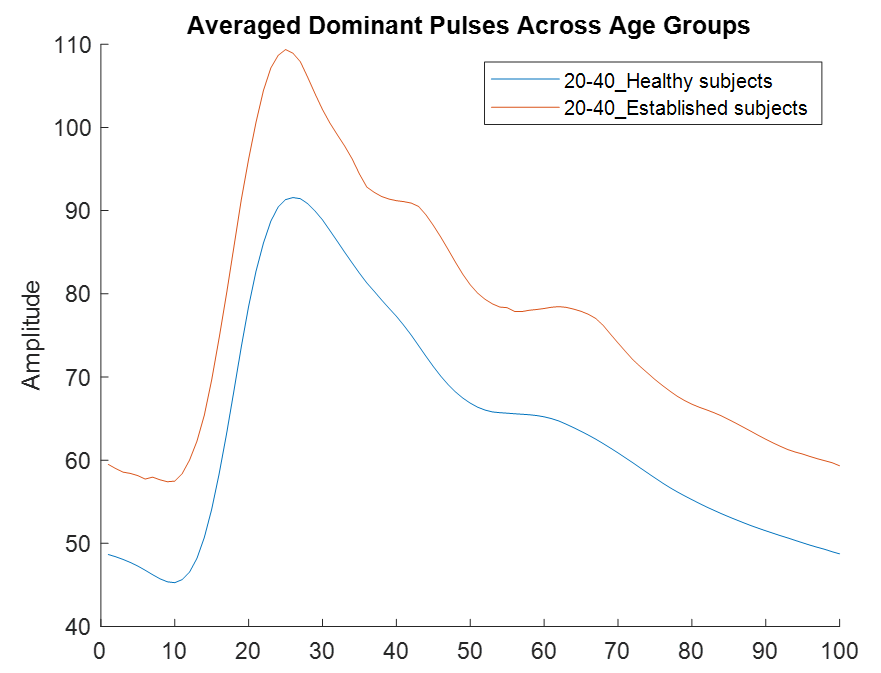}
        \caption{Averaged dominant pulse comparisons in age group [20,40] for healthy and established subjects.}
        \label{FigZhao8}
    \end{subfigure}

    \caption{Averaged dominant pulse comparisons in age group [20,40] for healthy subjects and different disease groups.}
    \label{FigZhao_DominantPulseComparison}
\end{figure*}

MOCAIP has the ability to extract the dominant pulse from the pulsatile signals. A dominant pulse in the MOCAIP toolbox is the representative pulse waveform selected from a group of pulses in a time window because it best reflects the stable, typical pulse morphology while reducing noise and artifacts. Here, we analyze and compare the averaged dominant pulse across various age groups for healthy subjects as shown in Fig. \ref{FigZhao5}. It has been observed that as age increases, the amplitude of the dominant pulse decreases with increased waveform complexity. This could be due to the vascular stiffening, wave reflection, and reduced cerebral compliance \cite{ChienVinuela2025}.  

Furthermore, the averaged dominant pulse comparisons across the same age group [20, 40] for healthy subjects and diseased subjects are shown from Fig. \ref{FigZhao6} to \ref{FigZhao8}. Fig. \ref{FigZhao6} shows the averaged dominant pulse comparisons in the age group [20, 40] for healthy and acute stroke subjects. Fig. \ref{FigZhao7} shows the comparisons between the healthy and post stroke subjects. Fig. \ref{FigZhao8} shows the comparisons between the healthy subjects and established subjects. Higher CBv amplitude and a greater number of waveform peaks were observed in the acute stroke and established groups. These waveform alterations may reflect abnormal cerebrovascular hemodynamics, as TCD-derived CBv waveforms have been shown to provide information related to cerebrovascular occlusion, stenosis, and pathological waveform morphology \cite{ThorpeThibeault2020}. Potential physiological mechanisms include impaired or heterogeneous cerebral autoregulation after stroke, compensatory changes in cerebral perfusion, and increased flow through collateral pathways \cite{MaGuo2018}. Moreover, vascular narrowing, disturbed or turbulent flow, and abnormal wave propagation near stenotic vessels may contribute to additional peaks in pulsatile CBv signals \cite{Kim2019}.

\subsubsection{Feature extraction}
\begin{figure}
    \centering
    \includegraphics[width=1.0\textwidth]{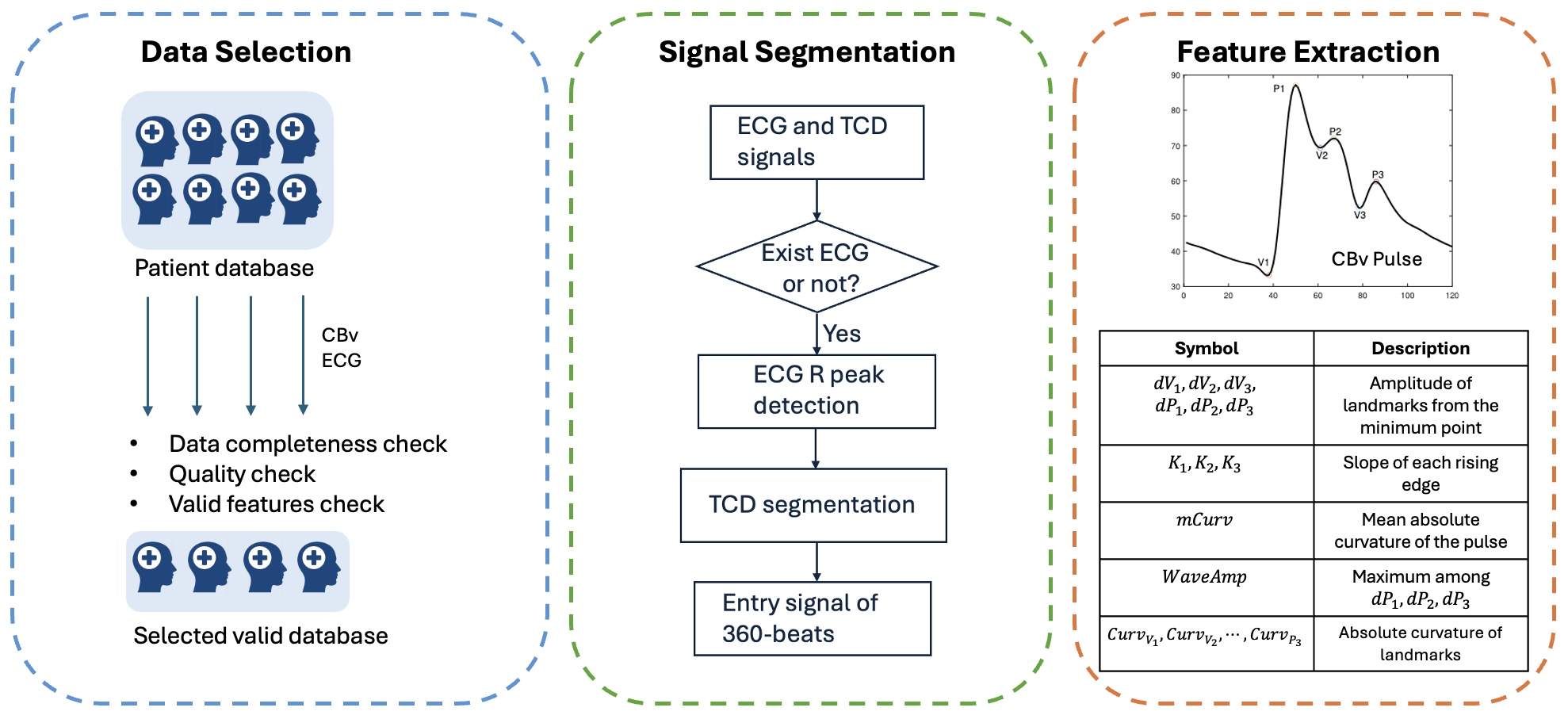}
    \caption{Data processing and feature extraction procedures for brain vascular age prediction using CBv. Left: Data selection procedure from the database. Middle: Signal segmentation for CBv pulse with and without ECG. Right: Feature extraction for CBv pulse using MOCAIP.}
    \label{FigZhao9}
\end{figure}

Fig. \ref{FigZhao9} shows the data processing in general, including the data selection, signal segmentation, and feature extraction. The 445 entries were selected using the following procedures: (1) data completeness and quality check, and (2) feature validity check. After the data completeness and quality check, signals from each recording were segmented into entries of 360 beats using the MOCAIP toolbox. Each recording is at least 360 beats long. Each entry was then manually reviewed by verifying the validity of the extracted MOCAIP features. There are 128 features in total, and some representative features are shown on the right-hand side of Fig.  \ref{FigZhao9}. All datasets, either non-invasive or invasive, are collected synchronously and resampled at 400 Hz. For entries with ECG, the R-R interval is computed using a general biomedical signal processing toolbox for QRS detection. The process of signal segmentation is shown in the middle diagram. The figure and table on the right-hand side show the feature extraction procedure for the MOCAIP toolbox. Other than the 128 features extracted from the MOCAIP toolbox, the body mass index (BMI) and heart rate variability features extracted from the ECG signal are adopted together as a feature vector for the machine learning algorithm. In total, there are 137 numerical features extracted from the TCD and ECG signals. There are various mathematical metrics that have been calculated in MOCAIP. The right-hand side of Fig. \ref{FigZhao9} shows some representative features, including the amplitude of landmarks from the minimum point, the slope of each rising edge, the mean absolute curvature of the pulse, and the absolute curvature of landmarks. After applying MOCAIP to the healthy and diseased subjects, the top 10 features with the largest group difference are obtained using the variations from the mean value of the features. The features with the largest group difference are computed following the procedures below:

Assume the original feature matrix is
\[
X =
\begin{bmatrix}
x_{11} & x_{12} & \cdots & x_{1p} \\
x_{21} & x_{22} & \cdots & x_{2p} \\
\vdots & \vdots & \ddots & \vdots \\
x_{N1} & x_{N2} & \cdots & x_{Np}
\end{bmatrix},
\]
where \(N\) is the total number of samples and \(p\) is the number of features.
The element \(x_{ij}\) represents the value of the \(j\)-th feature for the \(i\)-th sample.

If standard scaling is applied, each feature is normalized as
\[
z_{ij} = \frac{x_{ij} - \mu_j}{\sigma_j},
\]
where \(\mu_j\) and \(\sigma_j\) are the mean and standard deviation of the \(j\)-th feature, respectively. They are defined as
\[
\mu_j = \frac{1}{N}\sum_{i=1}^{N} x_{ij},
\]
and
\[
\sigma_j =
\sqrt{
\frac{1}{N-1}
\sum_{i=1}^{N}
\left(x_{ij}-\mu_j\right)^2
}.
\]

Therefore, the scaled feature matrix can be written as
\[
X_{\text{scaled}} =
\begin{bmatrix}
z_{11} & z_{12} & \cdots & z_{1p} \\
z_{21} & z_{22} & \cdots & z_{2p} \\
\vdots & \vdots & \ddots & \vdots \\
z_{N1} & z_{N2} & \cdots & z_{Np}
\end{bmatrix}.
\]

Let each sample belong to one of \(G\) groups. The group label of the \(i\)-th sample is denoted as
\[
g_i \in \{1,2,\ldots,G\}.
\]

For the \(g\)-th group and the \(j\)-th feature, the group-wise mean of the scaled feature values is computed as
\[
\bar{z}_{gj}
=
\frac{1}{n_g}
\sum_{i:g_i=g} z_{ij},
\]
where \(n_g\) is the number of samples in group \(g\).

The group-wise feature mean matrix is then given by
\[
M =
\begin{bmatrix}
\bar{z}_{11} & \bar{z}_{12} & \cdots & \bar{z}_{1p} \\
\bar{z}_{21} & \bar{z}_{22} & \cdots & \bar{z}_{2p} \\
\vdots & \vdots & \ddots & \vdots \\
\bar{z}_{G1} & \bar{z}_{G2} & \cdots & \bar{z}_{Gp}
\end{bmatrix}.
\]

To quantify how much each feature differs across groups, the variance of the group means is calculated for each feature. For the \(j\)-th feature, this variance is defined as
\[
V_j =
\frac{1}{G-1}
\sum_{g=1}^{G}
\left(
\bar{z}_{gj} - \bar{z}_{\cdot j}
\right)^2,
\]
where
\[
\bar{z}_{\cdot j}
=
\frac{1}{G}
\sum_{g=1}^{G}
\bar{z}_{gj}
\]
is the average of the group means for the \(j\)-th feature.

Finally, the features are ranked according to their variance values:
\[
V_{j_1} \geq V_{j_2} \geq \cdots \geq V_{j_p}.
\]

\begin{figure}
    \centering
    \includegraphics[width=0.85\textwidth]{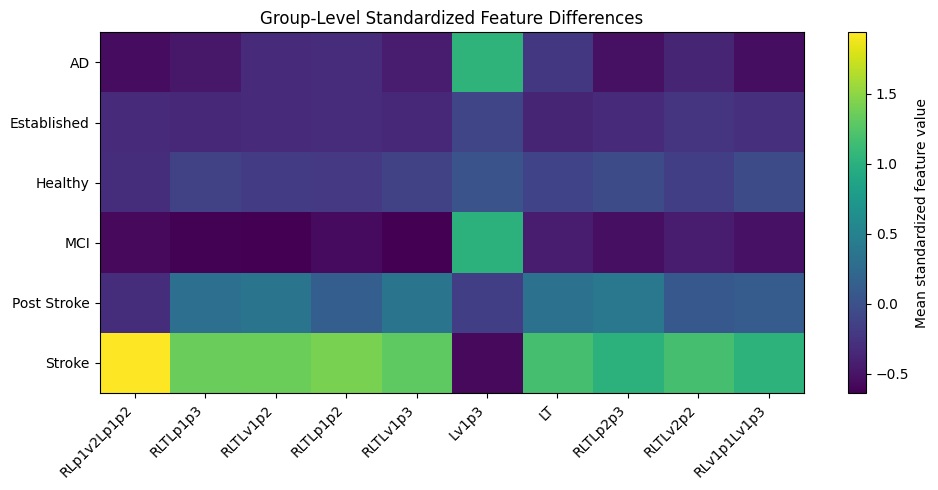}
    \caption{Group-level standardized feature differences for MOCAIP features extracted from the healthy and the diseased subjects.}
    \label{FigZhao10}
\end{figure}

The top \(K\) features are selected as
\[
\mathcal{F}_{\text{top}}
=
\{j_1, j_2, \ldots, j_K\},
\]
where \(K=10\) in this paper. Thus, the selected features are those with the largest variation in group-wise mean values across different groups. The top 10 features obtained are RLp1v2Lp1p2, RLTLp1p3, RLTLv1p2, RLTLp1p2, RLTLv1p3, Lv1p3, LT, RLTLp2p3, RLTLv2p2, and RLv1p1Lv1p3 as defined in MOCAIP. The mathematical definitions \cite{AsgariGonzalez2012} of the top 10 features can be found in Table \ref{Zhao.t1}. From Table \ref{Zhao.t1} we can see that there is a large group difference in terms of the ratio among time delays, time delay among landmarks, and the time delay between the peaks. A group-level visualization of the standardized features differences can be found in Fig. \ref{FigZhao10}. From Fig. \ref{FigZhao10}, it is observed that the acute stroke patients exhibit significant feature differences compared with the other subject groups. It is interesting to notice that for most of the features there are no large differences between the healthy subjects and the subjects with other diseases. There is minor difference in terms of the time delay among landmarks Lv1p3 for various subjects.

\begin{table*}[ht!] 
	\centering 
	\caption{Top 10 MOCAIP features with the largest group difference.}
    \footnotesize
	\begin{tabular}{c|ll} 
		\hline
		  Feature & Definition\\\hline\hline
		RLp1v2Lp1p2 & Ratio among time delays \\\hline
		  RLTLp1p3 &  Ratio among time delays \\\hline
        RLTLv1p2 & Ratio among time delays \\\hline
        RLTLp1p2 & Ratio among time delays \\\hline
        RLTLv1p3 & Ratio among time delays \\\hline
        Lv1p3 & Time delay among landmarks \\\hline
        LT & Time delay of $v_1$ to ECG QRS peak \\\hline
        RLTLp2p3 & Ratio among time delays \\\hline
        RLTLv2p2 & Ratio among time delays \\\hline
        RLv1p1Lv1p3 & Ratio among time delays \\\hline
	\end{tabular}
    \label{Zhao.t1}
\end{table*}

\subsection{Machine learning algorithms}
Various machine learning algorithms are adopted for the brain vascular age prediction, including the eXtreme Gradient Boosting (XGBoost), CatBoost, Random Forest, Gaussian process regressor, and the Tabular foundation model (TabPFN). 
XGBoost, CatBoost, and Random Forest are typical tree-based machine learning algorithms. Tree-based ensemble algorithms have been extensively applied in healthcare outcomes prediction, especially for structured clinical data such as EHRs, mortality prediction, readmission, and risk stratification \cite{DoupeFaghmous2019}. XGBoost has been applied for predicting 30-days mortality for MIMIC-III patients with sepsis-3 \cite{HouLi2020} and hospital readmission \cite{ZhaoYoo2021}. Multiple ensemble tree-based algorithms have been adopted for readmission evaluation for stroke patients in \cite{DarabiHosseinichimeh2021}. CatBoost has also been extensively applied for dementia risk prediction \cite{DingMandapati2023}, ICU mortality prediction \cite{SafaeiSafaei2022}, and carotid sonographic features analysis for the prediction of recurrent stroke \cite{LinLaw2022}. Random Forest is not as popular as XGBoost and CatBoost. However, it has been applied for dementia-related neuropsychiatric symptoms prediction \cite{MarGorostiza2020}, incident delirium \cite{CorradiThompson2018}, and intensive care unit readmission \cite{RojasCarey2018}. 

Tabular foundation model (TabPFN) \cite{HollmannMuller2025} is a specific foundation model trained using millions of synthetic tabular datasets. It outperforms the traditional algorithms, such as gradient-boosted decision trees, especially when dealing with datasets with a heterogeneity property. TabPFN has been extensively applied in healthcare \cite{LiYang2025, LuoYuan2025, TranByeon2024}. \cite{LuoYuan2025} proposed an integrated multimodal engine based on TabPFN for image data processing. Recently, TabPFN was also adopted in \cite{LiYang2025} for lymphvascular invasion prediction in invasive breast cancer. A hybrid LighGBM-TabPFN model was proposed in \cite{TranByeon2024} for dementia prediction in Parkinson's disease. TabPFN shows significant performance improvement when dealing with smaller tabular datasets and missing values. 
In this study, the XGBoost, CatBoost, Random Forest, Gaussian process regressor, and TabPFN are adopted for brain vascular age prediction. The MOCAIP and heart rate variability features extracted from the healthy subjects are adopted as inputs, and the age is adopted as output for the machine learning model. After the machine learning model is trained, it is applied to the diseased subjects for early brain disease detection. This approach is based on the assumption that the physiological features extracted from healthy and diseased subjects exhibit distinct characteristics. The healthy subjects are split into training and testing datasets. Furthermore, several resampling strategies were evaluated to address group imbalance; however, they did not consistently improve prediction performance. It has been observed that using healthy datasets above difference age threshold as training datasets yield different results. To guarantee the estimation accuracy, the final model was trained using the healthy subject distributed above age 50 with subgroup-level performance evaluation to assess potential bias across cohorts.

\section{Results}
\label{sec_results}
\begin{figure}
    \centering
    \includegraphics[width=1.0\textwidth]{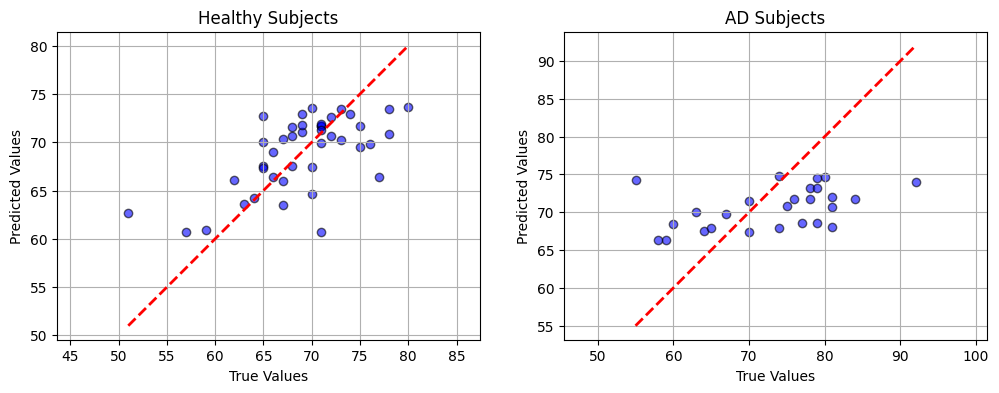}
    \caption{Brain vascular prediction results from healthy and AD subjects.}
    \label{FigZhao11}
\end{figure}

\begin{figure}
    \centering
    \includegraphics[width=1.0\textwidth]{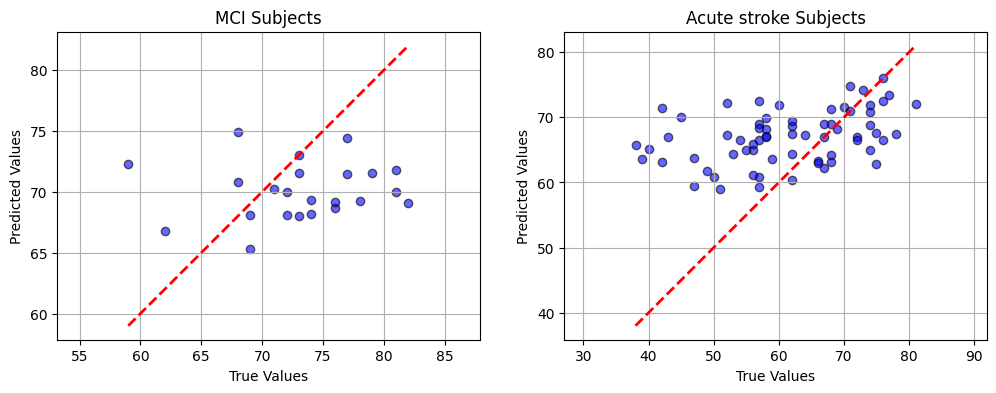}
    \caption{Brain vascular prediction results from subjects with MCI and acute stroke.}
    \label{FigZhao12}
\end{figure}

\begin{figure}
    \centering
    \includegraphics[width=1.0\textwidth]{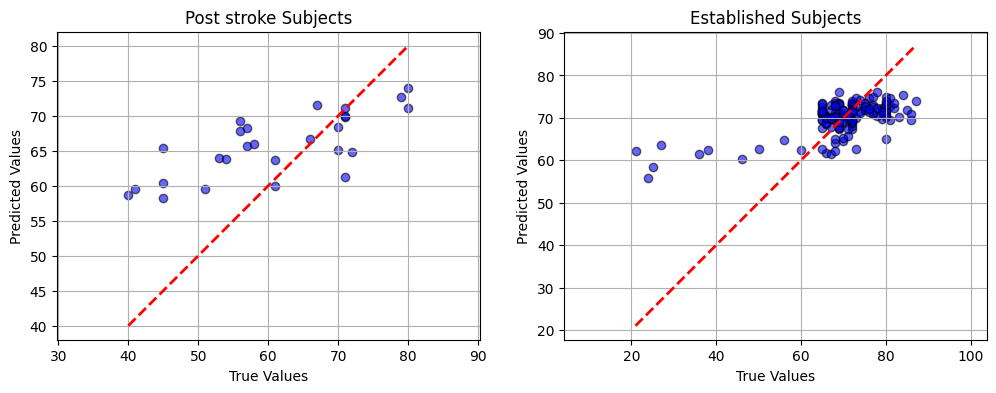}
    \caption{Brain vascular prediction results from post stroke and established subjects.}
    \label{FigZhao13}
\end{figure}

\begin{figure}
    \centering
    \includegraphics[width=0.8\textwidth]{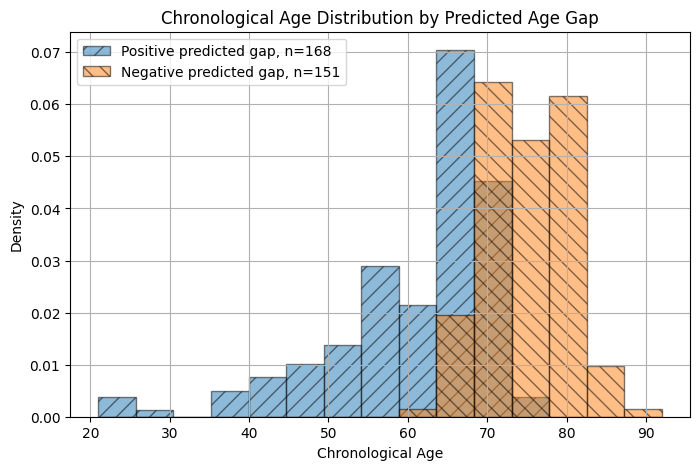}
    \caption{The positive and negative predicted brain vascular prediction gap for all subjects, including the healthy and diseased subjects.}
    \label{FigZhao14}
\end{figure}

\begin{figure}
    \centering
    \includegraphics[width=1.0\textwidth]{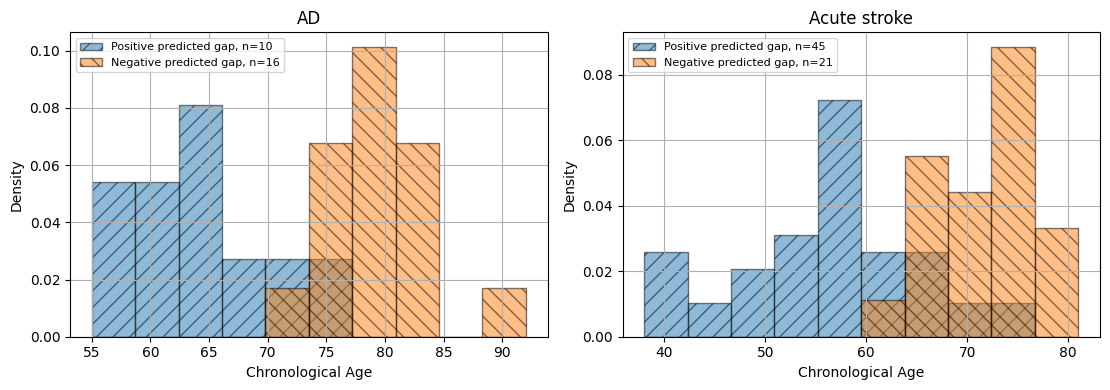}
    \caption{The positive and negative predicted brain vascular age gap for AD and Established subjects.}
    \label{FigZhao15}
\end{figure}

\begin{figure}
    \centering
    \includegraphics[width=1.0\textwidth]{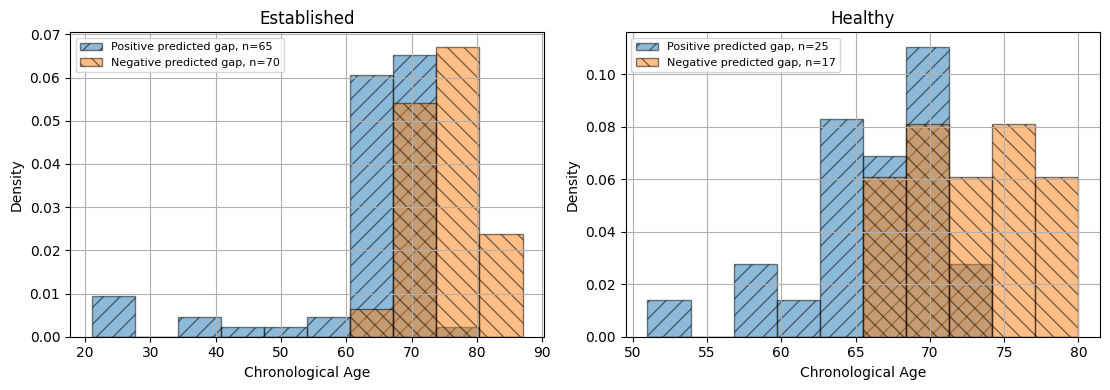}
    \caption{The positive and negative predicted brain vascular age gap for healthy and MCI subjects.}
    \label{FigZhao16}
\end{figure}

\begin{figure}
    \centering
    \includegraphics[width=1.0\textwidth]{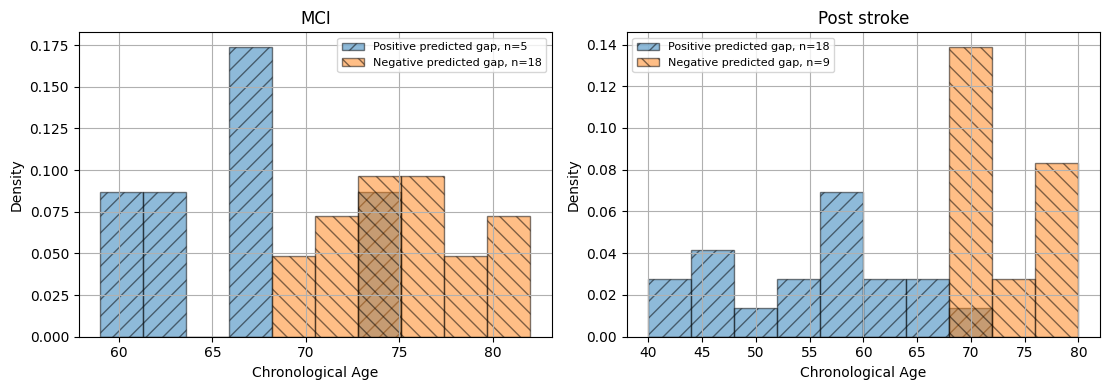}
    \caption{The positive and negative predicted brain vascular age gap for post stroke and stroke subjects.}
    \label{FigZhao17}
\end{figure}

The mean absolute errors of the predicted brain vascular age from healthy and diseased subjects for various algorithms are summarized in Table \ref{Zhao.t2}. From the results, we can see that the TabPFN yields the smallest mean absolute error of 3.36 for the healthy testing subjects. Moreover, the results from XGBoost, CatBoost, and Random Forest (RF) with balanced sample weighting \cite{PedregosaVaroquaux2011} are included in Table \ref{Zhao.t2}. It has been observed that there are minor improvements through incorporating the sample weighting algorithm. The XGBoost with sample reweighting yields the smallest mean absolute error for AD and MCI subjects. The RF with sample reweighting yields the smallest mean absolute error for post stroke and established subjects. Table \ref{Zhao.t3} shows the MAE difference between the healthy subjects and the diseased subjects. It is observed that the XGBoost shows the largest MAE difference between the healthy subjects and the subjects with stroke and established diseases. TabPFN shows the largest MAE difference between the healthy subjects and the subjects with MCI diseases. CatBoost with sample weighting shows the largest MAE difference between the healthy subjects and the subjects with AD and post stroke. On average, AD subjects showed 3.57 years of age acceleration. Patients with MCI showed 1.72 years of acceleration. Stroke patients exhibited the largest age acceleration, with 6.12 years. Post-stroke patients showed 5.51 years of acceleration, while established subjects showed 2.75 years of acceleration. The scatter plot for the predicted brain vascular age from healthy and diseased subjects is shown from Fig. \ref{FigZhao11} to Fig. \ref{FigZhao13}. Based on the results from Fig. \ref{FigZhao10}, we can see that there are no large feature differences between the subjects with various diseases, which is aligned with the mean absolute error shown in Table \ref{Zhao.t2}. The mean absolute errors from the trained machine learning models show no large difference between the subjects with AD, MCI, stroke, post stroke, and established diseases. Table \ref{Zhao.t4} shows the summary of the proportional, standard deviation, and mean of positive bran vascular gaps across various machine learning algorithms. The positive brain vascular age gap is defined as the predicted age minus the chronological age. From Table \ref{Zhao.t4} we can see that the stroke patients show the largest proportional, mean values, and standard deviation for positive brain vascular age gaps. Furthermore, the distribution of the positive predicted age gap and the negative predicted age gap is shown in Fig. \ref{FigZhao14}. From the figure we can see that there are 168 subjects show positive predicted gap and 151 subjects show negative predicted gap. Moreover, most of the positive predicted age gap is distributed in chronological age above 65. Most of the negative predicted age gap is distributed in chronological age below 65. Figs. \ref{FigZhao15} to \ref{FigZhao17} shows the chronological age distribution with positive predicted gap and negative predicted gap across different groups. It is observed that for different group, they are following the same distribution - most of the positive predicted age gap is distributed below age 65 $\sim$ 70 and most of the negative predicted age gap is distributed above age 70. This finding may suggest a tendency toward accelerated brain vascular aging among patients older than 65 years.

\begin{table*}[ht!] 
	\centering 
	\caption{Mean absolute error summary of brain vascular age prediction for healthy and diseased subjects from various machine learning algorithms. The minimum mean absolute error for each machine learning algorithm is highlighted in blue.}
    \footnotesize
	\begin{tabular}{c|lllllll} 
		\hline
		  Algorithms & Healthy & AD &  MCI & Acute stroke & Post stroke & Established\\\hline\hline
		XGBoost & 3.78 &  7.27   & 4.89  & 10.24 & 8.77 & 6.85 \\\hline
		  CatBoost & 3.70 & 7.51   & 5.78  & 10.00 & 10.11 & 6.46\\\hline
        Random forest & 3.56 & 7.24 & 5.53 & 9.77 & 9.82 & 6.37\\\hline
        XGBoost with sample weighting & 4.26 & \textcolor{blue}{7.03} & \textcolor{blue}{4.84} & 10.12 & 8.90 & 6.82\\\hline
        CatBoost with sample weighting & 3.69 & 7.51 & 5.78 & 10.00 & 10.11 & 6.46\\\hline
        RF with sample weighting & 3.51 & 7.19 & 5.30 & 9.63 & \textcolor{blue}{8.32} & \textcolor{blue}{6.07}\\\hline
        TabPFN & \textcolor{blue}{3.36} & 7.12 & 5.55 & \textcolor{blue}{8.92} & 8.39 & 6.08\\\hline\hline
        Mean & 3.69 & 7.27 & 5.38 & 9.81 & 9.21& 6.44\\\hline
        \multicolumn{7}{l}{\textit{Note}: RF denotes random forest.}
	\end{tabular}
    \label{Zhao.t2}
\end{table*}

\begin{table*}[ht!] 
	\centering 
	\caption{Mean absolute error difference summary of brain vascular age prediction between the healthy and diseased subjects from various machine learning algorithms.  The maximum mean error difference for each machine learning algorithm is highlighted in blue.}
    \footnotesize
	\begin{tabular}{c|lllll} 
		\hline
		  Algorithms & AD &  MCI & Acute stroke & Post stroke & Established\\\hline\hline
		XGBoost &  3.49   & 1.11  & \textcolor{blue}{6.46} & 4.99 & \textcolor{blue}{3.07}  \\\hline
		  CatBoost & 3.81   & 2.08  & 6.30 & 6.41 & 2.76\\\hline
        Random forest & 3.68 & 1.97 & 6.21 & 6.26 & 2.81\\\hline
        XGBoost with sample weighting & 2.77 & 0.84 & 5.86 & 4.64 & 2.56\\\hline
        CatBoost with sample weighting & \textcolor{blue}{3.82} & 2.09 & 6.31 & \textcolor{blue}{6.42} & 2.77\\\hline
        RF with sample weighting & 3.68 & 1.79 & 6.12 & 4.81 & 2.56\\\hline
        TabPFN & 3.76 & \textcolor{blue}{2.19} & 5.56 & 5.03 & 2.72\\\hline\hline
        Mean & 3.57 & 1.72 & 6.12 & 5.51 & 2.75\\\hline
	\end{tabular}
    \label{Zhao.t3}
\end{table*}

\begin{table*}[ht!] 
	\centering 
	\caption{Summary of the proportion, standard deviation, and mean of positive brain vascular age gaps from Tabular foundation model. The groups with the largest proportion, standard deviation, and mean of the brain vascular age gap are highlighted in blue.}
    \footnotesize
	\begin{tabular}{c|llllll} 
		\hline
		  Metrics & Healthy & AD &  MCI & Stroke & Post stroke & Established\\\hline\hline
		Proportion &  0.60  & 0.38  & 0.22 & \textcolor{blue}{0.68} & 0.67 & 0.48\\\hline
		  Mean & -0.01  & -2.51  & -3.13 & \textcolor{blue}{5.67} & 4.92 & 0.37\\\hline
        STD & 4.36 & 8.09 & 5.86 & \textcolor{blue}{9.98} & 9.17 & 9.17\\\hline
        \multicolumn{7}{l}{\textit{Note}: STD denotes standard deviation.}
	\end{tabular}
    \label{Zhao.t4}
\end{table*}

\clearpage

\section{Conclusions}
\label{sec_conclusion}
In this paper, the TCD signal is adopted for brain vascular age prediction. Mathematical features extracted using MOCAIP and heart rate variability toolbox are adopted as inputs for the machine learning algorithm. Various machine learning algorithms, including XGBoost, CatBoost, Random Forest, and Tabular foundation model, are adopted for brain vascular age prediction. It has been observed that the Tabular foundation model yields the smallest MAE for testing healthy subjects. XGBoost shows the largest MAE difference for stroke and established subjects, CatBoost shows the largest MAE difference for subjects with AD and post stroke. However, it has been observed that brain vascular age prediction depends heavily on the distribution of the age. In general, the age distribution for healthy subjects is highly imbalanced, and there are limited approaches existing for dealing with imbalanced datasets for brain age prediction. Although balanced sample weighting was adopted in this study to address the imbalanced age distribution, only limited performance improvement was observed. In the future, it is worth investigating a more effective resampling approach for brain vascular age prediction with imbalanced datasets. Furthermore, the unequal sample sizes among diseased groups could introduce bias into the machine learning model predictions. Demographic information, including gender and BMI, along with measurement-related factors such as estimation position, should be considered during the modeling and evaluation of machine learning models.

\bibliographystyle{unsrtnat}
\bibliography{sn-bibliography}

\end{document}